\documentclass{article}

\usepackage{arxiv}

\usepackage[utf8]{inputenc} 
\usepackage[T1]{fontenc}    
\usepackage{setspace}
\usepackage{hyperref}       
\usepackage{url}            
\usepackage{booktabs}       
\usepackage{amsfonts}       
\usepackage{nicefrac}       
\usepackage{microtype}      
\usepackage{lipsum}		
\usepackage{graphicx}
\usepackage{natbib}
\usepackage{doi}

\usepackage{amsmath}
\usepackage{tabularx}
\usepackage[format=plain, justification=justified]{caption}
\usepackage{subcaption}
\title{Heterogeneous Domain Adaptation and Equipment Matching: DANN-based Alignment with Cyclic Supervision (DBACS)}

\author{{\hspace{1mm}Natalie ~Gentner}\thanks{natalie.gentner@infineon.com (Natalie Gentner) } \\
\\
	Infineon Technologies AG\\
	Am Campeon 1-15, \\
 85579 Neubiberg, Germany \\
	\And
	{\hspace{1mm}Gian Antonio ~Stusto} \thanks{gianantonio.susto@unipd.it (Gian Antonio Susto)}\\
 \\
	Department of Information Engineering\\
	University of Padova\\
	Via Gradenigo 6/B, \\
 35131 Padova, Italy\\
}

\hypersetup{
pdftitle={Heterogeneous Domain Adaptation and Equipment Matching: DANN-based Alignment with Cyclic Supervision (DBACS)},
pdfsubject={cs.LG},
pdfauthor={Natalie ~Gentner, Gian Antonio ~Susto},
pdfkeywords={deep learning, equipment matching, heterogeneous domain adaptation, multi-view learning, semiconductor manufacturing, virtual metrology}
}

\begin{document}
\maketitle

\begin{abstract}
Process monitoring and control are essential in modern industries for ensuring high quality standards and optimizing production performance. These technologies have a long history of application in production and have had numerous positive impacts, but also hold great potential when integrated with Industry 4.0 and advanced machine learning, particularly deep learning, solutions. However, in order to implement these solutions in production and enable widespread adoption, the scalability and transferability of deep learning methods have become a focus of research. While transfer learning has proven successful in many cases, particularly with computer vision and homogenous data inputs, it can be challenging to apply to heterogeneous data. 

Motivated by the need to transfer and standardize established processes to different, non-identical environments and by the challenge of adapting to heterogeneous data representations, this work introduces the Domain Adaptation Neural Network with Cyclic Supervision (DBACS) approach. DBACS addresses the issue of model generalization through domain adaptation, specifically for heterogeneous data, and enables the transfer and scalability of deep learning-based statistical control methods in a general manner. Additionally, the cyclic interactions between the different parts of the model enable DBACS to not only adapt to the domains, but also match them. To the best of our knowledge, DBACS is the first deep learning approach to combine adaptation and matching for heterogeneous data settings. For comparison, this work also describes and analyzes subspace alignment and a multi-view learning method that deals with heterogeneous representations, called views, by mapping data into correlated latent feature spaces. Finally, the DBACS method, with its ability to adapt and match, is applied to a virtual metrology use case for an etching process run on different machine types in semiconductor manufacturing.
\end{abstract}


\section{Introduction}
Process control and monitoring are essential elements in any automated production setting. Both have a long history of use, particularly in specialized and demanding manufacturing environments. In recent years, the complexity of these systems has made them the focus of ongoing research, particularly in the context of Industry 4.0 and due to the increasing usage of sophisticated artificial intelligence-based solutions. While various machine learning and deep learning techniques have been applied successfully to a wide range of data types, the current focus is on scalability and the generalization of models, particularly for non-standardized environments.

Despite the potential of machine learning-based technologies to improve automation in production, there are several issues that continue to limit their widespread success. These include limited data availability, small data sets, lack of standardization, low or inconsistent data quality, and complex, fragmented data. These challenges can make it difficult to transfer and generalize models, hindering progress towards higher levels of fab automation and overall digitalization. As a result, the focus is now on standardization and scalability, particularly for application-driven research. This is important due to the financial and technological investments required for method and model development, as well as the need for 24/7 support for critical production infrastructure and maintenance in highly automated environments.

In non-standardized environments, there are two main approaches to supporting the scalability of methods and model transfer, as discussed in the semiconductor literature: (i) matching and (ii) transfer learning, with a focus on domain adaptation. The goal of matching is to harmonize environments and processes by using data and expert knowledge, with the aim of eliminating differences. Transfer learning, on the other hand, uses a purely data-driven approach to change the data representation (but not the data itself or any equipment or process properties) in order to bring corresponding data sets closer together, or in the best case, make them indistinguishable.

However, most machine learning-based transfer learning methods, which are driven by computer vision and naturally homogeneous data input, are not designed to handle heterogeneous data. While this is not a common issue when modeling tasks use images as the main input, it becomes a significant challenge in semiconductor scenarios where an established process must be transferred to a different, non-identical equipment due to availability or utilization. This raises the question of how to match non-identical equipment and use knowledge gained from one tool to optimize the same process on a different, non-identical tool in order to improve output quality.

To address the research gap related to heterogeneous domain adaptation (DA), this paper introduces an extended version of DBAM called DANN-based Alignment with Cyclic Supervision (DBACS). This method, which was previously applied to a homogeneous VM modeling task in previous studies \citet{gentner2020, gentner2021}, has the ability to map unpaired samples in their original feature spaces, enabling the functionality of matching. This capability allows DBACS to naturally enrich the existing method.

This contribution methodically extends the work presented in \citet{gentner2021} by demonstrating an extension suitable for heterogeneous domain adaptation (DA) and matching. The main contributions of the proposed DBACS method are as follows:
\begin{itemize}
    \item DBACS is able to handle high data complexity caused by heterogeneous systems in production and is applicable to various data types, such as time series data;
    \item DBACS is able to tackle both supervised and unsupervised adaptation for heterogeneous input data using the original input feature spaces;
    \item DBACS enables model scaling by allowing the use of a well-trained model for another data set with no assumption of the same feature representation, but only identical underlying physical information;
    \item DBACS ensures interpretability and comparability of all parts of the model and allows unpaired feature matching on top of the adaptation in both directions.
\end{itemize}
To evaluate the performance of DBACS in the context of heterogeneous data, the method is compared to selected benchmark models, including subspace alignment (SA) using principle component analysis (PCA) with and without correlation alignment (CORAL) and canonical correlation analysis (CCA), a method well known in the field of multi-view learning.

Virtual metrology (VM), a representative of standard process control mechanisms, is chosen as a real-world application showcase for this study. VM, also known as a soft sensor, is a statistical model that predicts inline wafer properties based on process information and sensor measurements. Since its introduction to the semiconductor industry in 2005 \cite{chen2005vm}, VM has a long research history and has benefited greatly from the adoption of new modeling techniques driven by Industry 4.0 and the use of artificial intelligence. In addition to being useful for predictive maintenance, fault detection and classification, and defect classification, VM is a key mechanism for direct/early fault detection and enabling quality improvements by increasing monitoring capacity, control through real-time process corrections in combination with a Run-to-Run system, and smart capacity usage by preparing input for smart sampling strategies and improved decision making.

The rest of the paper is organized into six more sections: Section 2 introduces related literature, Section 3 formalizes the problem and presents the main model DBACS and selected benchmarks. Section 4 gives details on virtual metrology, etching process, data, preprocessing and the experimental design, while in Section 5 implementation details including hyperparameter and architectures as well as results are reported. Finally in Section 6 DBACS suitability for matching is discussed. Section 7 closes with conclusive remarks and future research directions are envisioned.

\section{Literature and Background}\label{sec:literature}
In this section we summarize literature related to both relevant methodological approaches as well as application works. 




One of the main issue in adopting ML-based solutions in complex production is the need for scalability. With a large number of machines, products, and recipes (e.g., in semiconductor production), it is often infeasible to build ad-hoc analytics solutions for each scenario. In this context, scalability in learning frameworks is critical.  In this context, scalability in learning framework is of fundamental importance; one of this is \emph{matching}, which has Matching has a long history in non-standardized manufacturing environments and can be implemented using both classical methods \cite{chouichi2020chamber} and DL techniques \cite{applied2020}.

With the rise of Domain Adversarial networks (DANNs) \cite{ganin2015domainadversarial} and the related concept of \emph{domain adaptation}, a theory made to deal with occurrence of different data distributions for one modeling task, there has been a surge in the number of publications focusing on transfer learning for semiconductor applications  \cite{kang2017effectiveness}, \cite{tsutsui2019virtual} and \cite{9749610}. Domain adaptation also enables semi-supervised learning, as demonstrated in \cite{farahani2020novel} and \cite{li2020learning}. Unsupervised domain adaptation for semiconductor applications has also been explored using DANNs  \cite{SHIM2022}. A  variety of metrics and losses can be used to measure distribution distances in domain adaptation settings, such as maximum mean discrepancy (MMD) \cite{azamfar2020}. For a broader overview of domain adaptation, see \cite {wang2018deep} for a computer vision survey and \cite{courty2016optimal} for an example using optimal transport. Generative models have also been widely used in production-related research. For example, \cite{lu2019pix} presents a generative adversarial network (GAN)-inspired approach using pseudo labeling to address class imbalance in defect inspection in industrial settings. While the literature on homogeneous domain adaptation for semiconductor applications is growing, there is still a lack of research on heterogeneous domain adaptation for specific semiconductor use cases. However, literature from other industry sectors shows promising results for heterogeneous domain adaptation tasks, such as classification of heterogeneous information networks \cite{ijcai2020}, image-to-text transfer \cite{fang2022hetero, tsai2016hetero} and the combination of distribution alignment via subspace mapping with pseudo-labeling \cite{Alipour2022HeterogeneousDA}.

Another approach for handling heterogeneous data distributions is multi-view learning (MVL) \cite{perry2021mvlearn}. A systematic overview of MVL can be found in \cite{sun2013mvl} and in \cite{multiview}. There are few examples of MVL applied to fault detection and performance systems in manufacturing environments, such as \cite{CHEN2016cca} and \cite{yucca}, which use correlation and Canonical Correlation Analysis (CCA) \cite{cca2004} for fault detection and performance evaluation. A review of MVL in the deep learning (DL) context is provided by \cite{YAN2021106}, while a range of CCA approaches is discussed in \cite{Chapman2021}.

Metrology and its relationship to process control have been discussed in the literature, such as in early works such as \cite{chen2005vm} and \cite{su2007control}. While metrology is essential for quality and control, it can be costly in terms of productivity, which has led to the development of numerous virtual metrology (VM) approaches in the literature. VM is still an active research topic, with state-of-the-art methods like isolation forest being used in a decision-based model framework (e.g., \cite{chien2022decision}). VM tools are often developed based on data from fault detection and classification (FDC) systems, which are monitoring software used to overview different types of equipment in semiconductor manufacturing. FDC data typically consists of descriptive statistics computed from raw, time-dependent physical sensor measurements installed on the equipment, making the VM problem a classic tabular data regression task. Given the high dimensionality of FDC data, feature selection is an important step in the context of tabular data VM and has been widely discussed in the literature
; see \cite{saeys2007} for a general review of selection techniques and \cite{kang2016vmfeature} or \cite{lynn2009virtual} and \cite{9066890} for more sophisticated VM specific preprocessing and selection techniques. Other notable regression methods for VM prediction include those presented in \cite{lynn2009virtual}, \cite{susto2012least} and \cite{park2016virtual}. \cite{CHEN2020192} compares tree-based methods for VM modeling to other regression methods and neural networks. 

Another set of approaches in the VM literature aims to solve the regression task using time series data collected from equipment sensor data. These approaches include those presented in \cite{park2016virtual} and add \cite{susto2015pdm}, which introduced the Supervised Aggregative Feature Extraction framework for feature selection. DL-based approaches have also been successfully employed for modeling with time series input data, such as in \cite{lee2020recurrent, MAGGIPINTO2018, maggipinto2019deepvm} and \cite{lee2020recurrent}.

\section{Proposed Approaches}\label{section:methodology}
In this section, a general description and mathematical formalization of a modeling task with heterogeneous input are given, including the necessary assumptions. We also provide a formal description of the methods and algorithms used to solve the task under exam.

The modeling task at hand is formulated as regression, with the goal of scaling a selected model to make it usable for two data sets with different distributions and heterogeneous data representations. Since the input spaces do not have a common subspace sufficient for the task, the modeling must be done in a domain-specific manner. To address scalability, the goal is to use a trained statistical model for both data sets in parallel while minimizing the prediction error and maximizing the accuracy of a dedicated model. To achieve this, we compare methods from the field of domain adaptation and multi-view learning.

First, we mathematically formalize the regression task followed by selected methods. Let $f_S$ define a modeling task, let a hypothesis class $\mathcal{H}$ be a set of all possible modeling functions $h \in \mathcal{H}$ that are considered for a specific task. Let $X_S\subset \mathbb{R}^{n_S}$ be defined as the first input space and $Y$ as the output space. The output space $Y$ is defined as $Y\subset \mathbb{R}$ in case of a regression task. A distribution over $X_S \times Y$ is called source domain. For time series data, let $X_S\subset \mathcal{T}\times\mathbb{R}^{n_S}$ where $\mathcal{T}$ describes the set of considered points in time and $x^t_S \in \mathbb{R}^{n_S}$ a sample from the source feature space taken at a fixed point in time $t\in \mathcal{T}$. A learning algorithm is provided with a source data set $S$ drawn i.i.d. from the source domain $D_S$ with $X_S \times Y_S$, $X_S \subset X$, $Y_S \subset Y$. In the SSL setting, it is distinguished between labeled and unlabeled data and define $S:=SL \cup SU$ where $SU$ stands for the unlabeled source sample subset and $SL$ for the labeled one. Without loss of generality for UDA and SSL, $SU = \emptyset$ sine the source domain is assumed to be labeled. Hence
\begin{equation}
    S=\{X_{S},Y_{S}\}=\{X_{SL},Y_{SL}\}=\{x^i_S,y^i_S\}^{n_S}_{i=1} \sim \{D_S\}^{n_s},
\end{equation}
with $n_S$ being the number of drawn samples (all labeled) and therefore $X_S=X_{SL} \subset \mathbb{R}^{n_S}$, $Y_S=Y_{SL} \subset Y$.

A learning algorithm is provided with a second set $T$ drawn i.i.d. from the target domain $D_T$ with different data distribution, representation and feature space. Hence, let $T=TL \cup TU$ be the second called target data set $T$ drawn i.i.d. from a target domain $D_T$ with a distribution over $X_T \times Y_T$, $X_T \subset \mathbb{R}^{n_T}$, $Y_T \subset Y$, and consisting of unlabeled $TU$ and/or labeled $TL$ samples.
\begin{equation}
    TL =\{X_{TL},Y_{TL}\}=\{x^j_T,y^j_T\}^{n_T-l}_{j=1} \sim \{D_T\}^{n_t-l};
    \end{equation}
\begin{equation}
    TU =\{X_{TU}\}=\{x^j_T\}^{n_T}_{j=n_T-l+1} \sim \{D^X_T\}^{n_t};
\end{equation}
with $n_T$ being the number of drawn target samples, therefore $X_{TL} \subset X_T \subset \mathbb{R}^{n_T}$, $Y_{TL} \subset Y_T \subset Y$ and $X_{TU} \subset X_T \subset \mathbb{R}^{n_T}$. For time series data, let $X_T\subset \mathcal{T}\times\mathbb{R}^{n_T}$ where $\mathcal{T}$ describes the set of considered points in time and $x^t_T \in \mathbb{R}^{n_T}$ a sample from the target feature space taken at a fixed point in time $t\in \mathcal{T}$.

\subsection{DANN-based Alignment with Cyclic Supervision (DBACS)}
\begin{figure}[ht]
\centering
\includegraphics[width=0.8\textwidth]{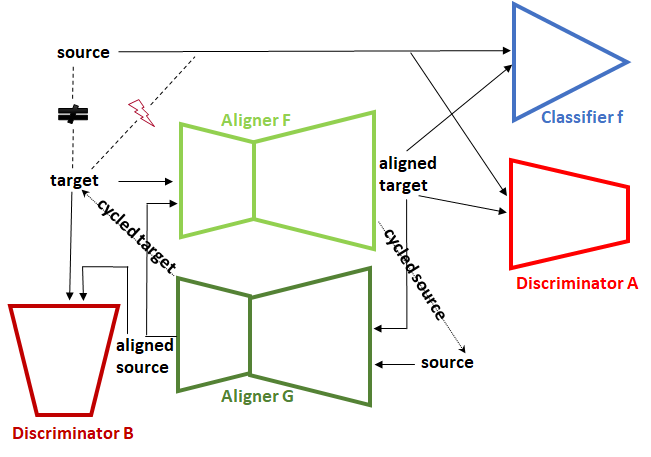}
\caption{\textbf{Graphical representation of the proposed DBACS system} exploiting input data from two non-identical domains. The arrows represent the data flows. An autoencoder shaped aligner can be used for noise reduction especially for homogeneous DA but is not mandatory.}\label{fig: dbamext}
\end{figure}

In this work, we present a new framework, called DANN-based Alignment with Cyclic Supervision (DBACS). The DBACS approach (illustrated in Figure \ref{fig: dbamext}) is an extented version of DBAM \cite{gentner2021} and is designed for binary heterogeneous domain adaptation using source and target domain. DBACS consists of five main parts:
\begin{itemize}
    \item the baseline or reference \emph{prediction} model $P$;
    \item an encoder/alignment model called \emph{aligner} $F$ used to map the target domain to the source domain (the output of the aligner is called aligned);
    \item $F$ is connected to a second encoder/alignment model \emph{aligner} $G$ that maps the source domain to the target domain. By combining both aligners, it is possible to introduce \emph{cycle-consistency} by comparing source samples with its cycled sample and target samples with its cycled samples;
    \item a domain \emph{discriminator} $A$ for classification of source domain versus aligned target domain;
    \item Adversarial training in both directions is enabled by adding a second domain discriminator (called \emph{discriminator} $B$) for target versus aligned source comparison.
\end{itemize}
The various components of the DBACS architecture are discussed in the following.

\paragraph{Prediction loss}
Let $h_P: X_S \rightarrow Y$ be a dedicated statistical model trained on a data set $S$, let $S$ be a labeled source sample set drawn i.i.d. from a domain $D_S$ with $n_S=|S|$ being the number of drawn samples. The neural network representation is parameterized by $\theta_P$ and $P(x_S,\theta_P)$ where $P$ is the model function with parameters $\theta_P$ that outputs the prediction for $x_S\in X_S$. The loss $L_P$ used for training and minimization is defined as:
\begin{align}
\min_{h_P\in\mathcal{H}}L_{P}(X_S)= \min_{h_P\in\mathcal{H}}L_{D_S}(h_P(X_S)) = \min_{\theta_P}L_{D_S}(X_S,\theta_P)
=\min_{\theta_P} \mathbf{L}_{\left(x,y\right) \in S}\left(P\left(x,\theta_P\right),y\right).
\end{align}
where $\textbf{L}$ is selected based on the modeling task at hand; for VM regression task we choose mean absolute error (MAE).

\paragraph{Cycle consistency loss}
Let  $h_F: X_T \rightarrow X_S$ be a statistical model function aligning target to source and $F(x_T,\theta_F)$ be its parameterized representation where $F$ is the model function with parameters $\theta_F$ that outputs the prediction for $x_T\in X_T$. Let  $h_G: X_S \rightarrow X_T$ be a statistical model function aligning source to target and $G(x_S,\theta_G)$ be its parameterized representation where $G$ is the model function with parameters $\theta_G$ that outputs the prediction for $x_S\in X_S$. Let $x_S \sim D_S$ the data distribution according to $D_S$ and $x_T \sim D_T$ according to $D_T$. Then, the cycle-consistency loss is defined as:
\begin{equation}
       L_{cycle_S}(X_S) := L_{G,F,D_S}(X_S) = \mathbf{L}_{x_S \sim D_S}\left(F(G\left(x_S,\theta_G),\theta_F\right)\right) = \mathbf{L}_{x_S \sim D_S} (F(G(x_S)),x_S);
\end{equation}
\begin{equation}
       L_{cycle_T}(X_T) := L_{F,G,D_T}(X_T) = \mathbf{L}_{x_T \sim D_T}\left(G(F\left(x_T,\theta_F),\theta_G\right)\right) = \mathbf{L}_{x_T \sim D_T} (G(F(x_T)),x_T).
\end{equation}
To give an example we follow the description in \cite{zhu2017} where the $L_1$ norm is used as cycle loss function:
\begin{align}
       L_{cycle_S}(X_S)=\mathbf{L}_{x_S \sim D_S} (F(G(x_S)),x_S)= \mathbf{E}_{x_S \sim D_S}\left[\left\|F(G(x_S)) - x_S \right\|_1\right] \\
       L_{cycle_T}(X_T)=\mathbf{L}_{x_T \sim D_T} (G(F(x_T)),x_T)= \mathbf{E}_{x_T \sim D_T}\left[\left\|G(F(x_T)) - x_T \right\|_1\right] 
\end{align}
In short, the cycle consistency loss is defined as
\begin{equation} 
    \mathcal{L}_{cyc}(F, G, X_S, X_T) = L_{cycle_S}(X_S) + L_{cycle_T}(X_T).
\end{equation}
Here, the optimization goal is to reproduce a bijective mapping so that each $x_S\in X_S$ is mapped to $X_T$ and back to $X_S$ with $F(G(x_S))\approx x_S$. The same goes for $x_T\in X_T$ with $G(F(x_T))\approx x_T$.

\paragraph{Adversarial loss}
Let $h_{D_A}: X_S \to I$, $h_{D_B}: X_T \to I$ with $I=[0,1]$ be two statistical model functions describing respectively the distance of source versus aligned target and target versus aligned source. Let $h_{D_A}$ be parameterized by $\theta_{D_A}$ and let $D_A(x_S,\theta_{D_A})$ be the parameter representation of discriminator A where $D_A$ is the model function with parameters $\theta_{D_A}$ that outputs the prediction for $x_S\in X_S$. Let $h_{D_B}$ be parameterized by $\theta_{D_B}$ and let $D_B(x_T,\theta_{D_B})$ be the parameter representation of discriminator B where $D_B$ is the model function with parameters $\theta_{D_B}$ that outputs the prediction for $x_T\in X_T$. Then the adversarial loss first for source $L_{adv_{S}}$ and second for target $L_{adv_{T}}$ is defined based on a selected loss function $\mathbf{L}$: 
\begin{align}
    L_{adv_{S}}(X_S,X_T) &:= L_{D_A,D_S}(X_S) - L_{D_A,D_T}(X_T) \nonumber\\&= \mathbf{L}_{x_S \sim D^X_S}\left(h_{D_A}\left(x_S\right)\right) -  \mathbf{L}_{x_T \sim D^X_T}\left(h_{D_A}\left(h_F(x_T)\right)\right) \nonumber \\
    &=\mathbf{L}_{x_S \sim D^X_S}\left(D_A\left(x_S,\theta_{D_A}\right)\right) - \mathbf{L}_{x_T \sim D^X_T}\left(D_A\left(F(x_T, \theta_{F}),\theta_{D_A}\right)\right)
\end{align}
\begin{align}
    L_{adv_{T}}(X_S,X_T) &:= L_{D_B,D_T}(X_T) - L_{D_B,D_S}(X_S) \nonumber\\&= \mathbf{L}_{x_T \sim D^X_T}\left(h_{D_B}\left(x_T\right)\right) -  \mathbf{L}_{x_S \sim D^X_S}\left(h_{D_B}\left(h_G(x_S)\right)\right) \nonumber \\
    &=\mathbf{L}_{x_T \in D^X_T}\left(D_B\left(x_T,\theta_{D_B}\right)\right) - \mathbf{L}_{x_S \sim D^X_S}\left(D_B\left(G(x_S, \theta_{G}),\theta_{D_B}\right)\right)
\end{align}
The adversarial loss is applied to the output of each discriminator and enables an adversarial training approach (see \cite{gentner2021}, \cite{gulrajani2017ganimproved}). For a regression modeling task: Let $I \subset \mathbb{R}$ or $I=\mathbb{R}$ and the discriminator a regression model with linear output activation function. Then the adversarial loss is defined as
\begin{equation}
    L_{adv_{S}}(X_S,X_T) = \mathbf{E}_{x_S \sim D_S}\left[D_A\left(x_S,\theta_{D_A}\right)\right] - \mathbb{E}_{x_T \sim D_T}[(D_A(F(x_T,\theta_F),\theta_{D_A}))],
\end{equation}
\begin{equation}
    L_{adv_{T}}(X_S,X_T) = \mathbf{E}_{x_T \sim D_T}\left[D_B\left(x_T,\theta_{D_B}\right)\right] - \mathbb{E}_{x_S \sim D_S}[(D_B(G(x_S,\theta_G),\theta_{D_B}))],
\end{equation}
where $\mathbf{E}$ defines the expected value and the loss is an approximation of the Wasserstein distance of two sampled distributions, for details see \cite{gulrajani2017ganimproved}.

\paragraph{Remark} It is recommended by \cite{zhu2017} based on \cite{taigman2017} to add one more additional loss term namely the identity loss. The idea is that if almost identical samples in the other domain occur, the aligner should perform close to an identity function. Since source and target are heterogeneous in our case we do not apply this kind of loss.\\

The training itself happens in an adversarial setting with a two-player game approach. The adversarial training routine includes parallel training of both aligner and both discriminator using adversarial loss plus inclusion of the additional loss terms. For training, two fixed training data sets $S$ and $T$ (training samples are drawn i.i.d from $D_S$ and $D_T$) are used. During the aligners training phase, the adversarial loss is minimized, during discriminator training phase it is maximized (or its negative value minimized).: 

\begin{itemize}
    \item The first competitor of the adversarial training is the discriminator $D_A$ trained to distinguish between source and aligned target data meaning optimizing the adversarial source loss. In parallel the discriminator $D_B$ is trained to distinguish between target and aligned source data also meaning optimizing the adversarial target loss. The optimization of the discriminator A and discriminator B loss $L_{D_{total}}$ is defined as
\begin{align}
   \max_{\theta_{D_A},\theta_{D_B}} L_{D_{total}}(X_S,X_T) &= \max_{\theta_{D_A}}L_{adv_S}(X_S,X_T) + \max_{\theta_{D_B}}L_{adv_T}(X_S,X_T)\nonumber\\
   &= \max_{\theta_{D_A}} L_{D_A,D_S}(X_S,\theta_{D_A}) - L_{D_A,D_T}(F(X_T,\theta_F),\theta_{D_A}) \nonumber \\
   &+ \max_{\theta_{D_B}} L_{D_B,D_T}(X_T,\theta_{D_B}) - L_{D_B,D_S}(G(X_S,\theta_G),\theta_{D_B})\nonumber\\
   &=\max_{\theta_{D_A}} \mathbf{L}_{x_S \in S}\left(D_A\left(x_S,\theta_{D_A}\right)\right) - \mathbf{L}_{x_T \in T}\left(D_A\left(F(x_T, \theta_F),\theta_{D_A}\right)\right) \nonumber\\
   &+ \max_{\theta_{D_B}} \mathbf{L}_{x_T \in T}\left(D_B\left(x_T,\theta_{D_B}\right)\right) - \mathbf{L}_{x_S \in S}\left(D_B\left(G(x_S, \theta_G),\theta_{D_B}\right)\right)\label{equ:dloss2}
\end{align}
\item The second competitor in the adversarial training is the aligner cycle. We define $L_{A_{total}}$ using the adversarial loss for both aligners and the cycle consistency loss. In case of labeled target data the aligner $F$ is also updated in order to optimize prediction loss $L_P$ for aligned target data. The adversarial part of the aligner losses is set in opposite direction compared to the ones used to update the two discriminator:
\begin{align}
 \min_{\theta_{F},\theta_{G}}L_{A_{total}}(X_S,X_T) & = \min_{\theta_{F},\theta_{G}}\lambda_{adv_S} L_{adv_S}\left(X_S,X_T\right) +  \lambda_P L_P\left(F(X_T)\right)
+ \lambda_{adv_T} L_{adv_T}\left(X_S,X_T\right) \nonumber \\
&= \min_{\theta_F,\theta_{G}} [- \mathbf{L}_{x_T \in T}\left(D_A\left(F(x_T, \theta_F),\theta_{D_A}\right)\right) \nonumber \\ &+ \lambda_P \mathbf{L}_{(x_T,y) \in TL}(P\left(F(x_T,\theta_F),\theta_P\right)) -\mathbf{L}_{x_S \in S}\left(D_B\left(G(x_S, \theta_G),\theta_{D_B}\right)\right)]
\end{align}
where  $\lambda_{(\cdot)}$ represents the weight assigned to each corresponding loss term. For $\lambda_P=0$ the training happens in an unsupervised setting where no target labels are available. A gradient penalty regularization term is added when updating both aligners following the recommendations of \cite{gulrajani2017ganimproved}.
\end{itemize}

\subsection{Subspace Alignment using Principle Component Analysis}
Subspace Alignment (SA), presented by \cite{basura2013sa}, linearly aligns subspaces generated by Principle Component Analysis (PCA).

SA was introduced as unsupervised DA method for classification task. Overall benefits of SA lies in the simplicity and in the speed of the method while still presenting high accuracy. For heterogeneous domain adaptation, we slightly adapt here the SA approach by first applying PCA separately to source and target and then align the corresponding subspaces using CORrelation ALignment (CORAL). CORAL by \cite{sun2016return} is an unsupervised domain adaptation method that aligns second order statistics of source and target domain. 

\paragraph{PCA} Principle Component Analysis (PCA) is a linear transformation of a vector space with respect to its points/vectors. The projection is created in a way that highest occurring variance is represented by the first latent dimension (the so-called first principle component), the second highest variance by the second principle component and so on. 
Let $X$ be a vector space, let $\psi: X \to X'$ define the nonlinear principle component transformation to be computed. Then PCA is formalized via 
\begin{equation}
    x' = \psi(x) = \Gamma^T x
\end{equation}
where $x' \in X'$ describes the transformed input, $\Gamma$ consists of the eigenvectors and is computed via $\Lambda = \Gamma^T\Sigma\Gamma$ where $\Lambda$ is a diagonal matrix defined by the eigenvalues and $\Sigma$ is the covariance matrix. PCA is applied to $X_S$ and $X_T$ accordingly resulting in $S'$ and $T'$ as projected input sets. Since PCA is a very well-known method, we refer to \cite{pca} for a more detailed description.

\paragraph{CORAL} Let $S' = \{x'_{S_i}\}$, $T' = \{x'_{T_i}\} $ be the PCA projected input sets from the source and target domains. Let $\Upsilon: X'_S \to X'_T$ with $\Upsilon(X'_S )=X'_{S}*A$ describe the feature transformation of the source space to the target space. Let $\mu_{S'},\mu_{T'}$ be the feature mean of $S'$, $T'$ and $C_{S'},C_{T'}$ the corresponding covariance matrices. Then, the distance between the covariance matrices (assuming normalized features with zero mean) is minimized by:
\begin{equation*}
    \min_A \left\| C_{\hat{S'}} - C_{T'} \right\|_F^2 = \min_A \left\| A^TC_{S'}A - C_T' \right\|_F^2
\end{equation*}
where $A$ is the matrix used in linear transformation that is applied to the source, $C_{\hat{S'}}$ describes the covariance of the transformed source features $S'^*A$ and $\left\| \cdot \right\|_F^2$ denoting the squared Frobenius norm selected as distance metric. It is called CORAL loss. In order to solve this equation, we follow Algorithm 1 in \cite{sun2016return} and compute first the covariance matrices followed by whitening the source and then recoloring it with the target covariance.

\subsection{Canonical Correlation Analysis (CCA)}
Canonical Correlation Analysis (CCA) defines linear transformation for each set of variables such that after the transformation the projected features are maximal correlated. A summary of the descriptions is taken from \cite{cca2004}.

\begin{figure}[htb]
\centering
\includegraphics[width=0.5\textwidth]{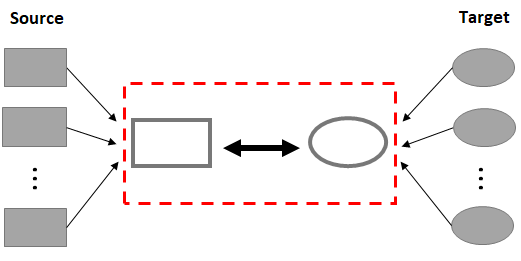}
\caption{\textbf{Visualization of Canonical Correlation Analysis (CCA).} The canonical components of source and target are a weighted combination of corresponding input features. The correlation of the canonical components within the red box is maximized. Similarly to PCA, the number of canonical components can be tuned.}\label{fig: cca}
\end{figure}

Let $S=\{x_{S}\}, T=\{x_{T}\}$ be two sample sets wanted to be projected into direction $w_S, w_T$. Let $\Phi_S: X_S \to X_S'$, $\Phi_T: X_T \to X_T'$ define the linear transformation for each domain. Then:
\begin{align}
\Phi_S(S)=S^{'}=S_{x_S,w_S}= \langle w_S,x_S \rangle, \nonumber \\ \Phi_T(T)=T^{'} = T_{x_T,w_T}= \langle w_T,x_T \rangle.
\end{align}
Specifically, it is looked for $w_S, w_T$ such that the correlation between the projected vectors is maximised, hence:
\begin{equation}
\rho= \max_{w_S, w_T} corr\left(S_{x_S,w_S},  T_{x_T,w_T}\right) = \max_{w_S, w_T} \frac{ \langle S_{x_S,w_S},T_{x_T,w_T} \rangle}{\|S_{x_S,w_S}\| \cdot \|T_{x_T,w_T}\|}.
\end{equation}
The previous equation can be reformulated as
\begin{equation}
\rho= \max_{w_S, w_T} \frac{w_s'\mathbb{E}[x_Sx_T']w_T}{\sqrt{w_S'\mathbb{E}[x_Sx_S']w_S w_T'\mathbb{E}[x_Tx_T']w_T}}
\end{equation}
with $\mathbb{E}$ denoting the discrete empirical expectation, $'$ denotes the transpose of a vector or a matrix and properties of the inner product are used.
Using the covariance matrix with 
\begin{equation} C= C(x_S,x_T)= \mathbb{E}[x_Sx_T]=\begin{bmatrix}
       C_{x_Sx_S} & C_{x_Tx_S} \\[0.3em]
       C_{x_Sx_T} & C_{x_Tx_T} 
     \end{bmatrix} 
\end{equation}
where C is a block matrix with the within-covariance $C_{x_Sx_S}, C_{x_Tx_T}$ and between-covariance matrices $C_{x_Sx_T}, C_{x_Tx_S}$ as entries. Finally the optimization problem can be formulated in the following way:
\begin{equation}
\rho= \max_{w_S, w_T} \frac{w_s'C_{x_Sx_T}w_T}{\sqrt{w_S'C_{x_Sx_S}w_S w_T'C_{x_Tx_T}w_T}}
\end{equation}
By checking that rescaling of $w_S, w_T$ does not change the problem, it can be maximized subject to 
\begin{align} 
w_S'C_{x_Sx_S}w_S = 1,\nonumber \\
w_T'C_{x_Tx_T}w_T = 1.
\end{align}
The formulation of the dual problem is used, hence computing the corresponding Lagrangian L leads to
\begin{equation}
L(\lambda,w_S, w_T) = w_s'C_{x_Sx_T}w_T - \frac{\lambda_S}{2}(w_S'C_{x_Sx_S}w_S - 1) - \frac{\lambda_T}{2}(w_T'C_{x_Tx_T}w_T - 1).
\end{equation}
The partial derivatives in the direction of $w_S, w_T$ are:
\begin{align} 
\frac{\partial L}{w_S} = C_{x_Sx_T}w_T - \lambda_SC_{x_Sx_S}w_S = 0,\label{eq:1}\\
\frac{\partial L}{w_T} = C_{x_Tx_S}w_S - \lambda_TC_{x_Tx_T}w_T = 0.\label{eq:2}
\end{align}
Multiplying (\ref{eq:2}) with $w_S*$ and multiplying (\ref{eq:1}) with $w_T*$ and subtracting the one from the other, define $\lambda=\lambda_S=\lambda_T$, assuming $C_{x_Tx_T}$ is invertible, rearrange the equation and use the partial derivative leaves to 
\begin{equation}
C_{x_Sx_T}C_{x_Tx_T}^{-1}C_{x_Tx_S}w_S = \lambda^2 C_{x_Sx_S}w_S
\end{equation}
which is equivalent to a generalised eigenproblem of the form $Ax = \lambda B x$. Using Cholesky decomposition, the previous can be even more simplified to a symmetric eigenvalue problem $Ax = \lambda x$. For visualization see Figure \ref{fig: cca}.

\section{Case Study: Dataset Description and Experimental Settings}
\subsection{Semiconductor Manufacturing: Etching process and Virtual Metrology}

Wafers are the basis for every semiconductor manufacturing process. A wafer consists of pure (99.9999\%) silicon, has a disc shape and houses several thousand chips (the end product) on average. The specific technology structure of a chip is built up layer by layer on the wafer during a couple of hundred process steps. Each wafer is considered a separate sample in this work.

Etching is a common process in semiconductor manufacturing and is frequently studied and discussed in semiconductor research and literature, along with chemical vapor deposition and implantation. The etching process removes material from a surface or transfers a structure created during the lithography step to the layer below \cite{hilleringmann, spanos2006}. Reactive-ion etching uses a high-frequency alternating energy field applied to the cathode on which the wafer is placed. Positively charged ions in the plasma are accelerated towards the wafer and collide with its surface at high kinetic energy, causing atoms from the wafer's surface to be dislodged from the crystal lattice, resulting in partial physical etching. In addition, a partial chemical reaction occurs due to the highly reactive free radicals.


The plasma etching process includes up to ten sub-steps during which input sensors must be adjusted to the target values specified in a recipe. Sensors that measure properties such as chamber pressure, applied high frequency voltage, gas type, gas flow, and wafer temperature, as well as electrode temperature and bias, play a crucial role in achieving the desired wafer properties. End point detection, or the etching time, is one of the most critical aspects of the process, as it is highly sensitive and closely related to other variables such as gases, pressure, current, and temperature. Incorrect etching times, inadequate end point detection, uncontrolled reactions, and interference in the chamber can negatively affect the layer thickness and overall quality and functionality of the wafer.

Process monitoring and control are essential for reliable, standardized, and repeatable production processes that produce high-quality products. In this work, we focus on a process control method called virtual metrology (VM) and analyze it through a case study involving an etching process. In general, control quantities are typically measured in metrology stations or tools after the process is completed, using multiple measurements on a sample of wafers. Traditional metrology is a univariate or multivariate control system that uses control charts with defined upper and lower control limits to monitor process performance. However, due to cost and time constraints, not all wafers can be physically measured after the process.

Virtual metrology (VM) or soft sensing modules utilize data collected by process equipment to model the relationship between wafer properties and process input and feedback sensor measurements. VM techniques allow for the inclusion of non-measured but predicted control measures in order to enhance analysis. VM technologies offer several benefits, including:
\begin{itemize}
    \item \emph{costs and time savings} due to reduced mandatory measurements;
    \item \emph{quality assurance} through enhanced and comprehensive monitoring;
    \item \emph{real-time control, assessment and process updates} in conjunction with Run-to-Run controllers \cite{su2007control};
    \item \emph{data-driven process optimization} including fault detection, root cause analysis and improved sample selection \cite{feng2019online}.
\end{itemize}

\subsection{Data Preparation}

\begin{figure}[htb]
\centering
\includegraphics[width=0.8\textwidth]{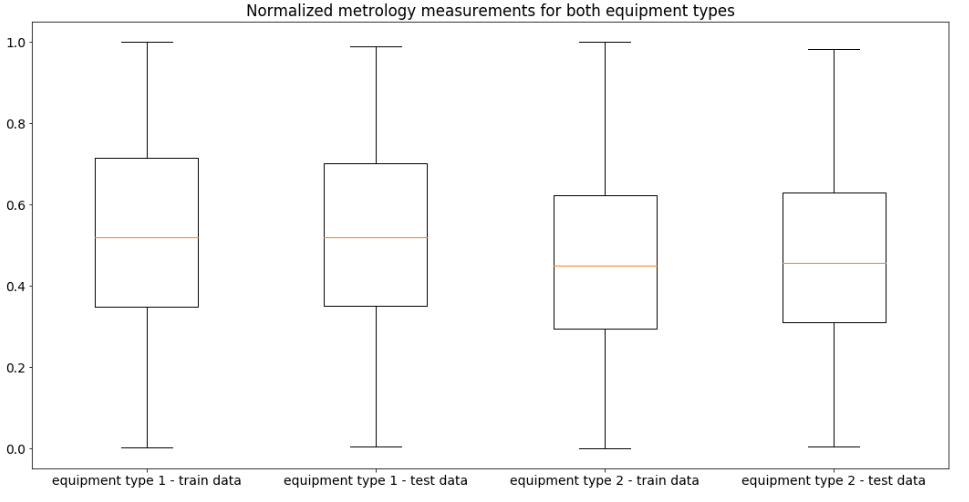}
\caption{\textbf{Boxplot of normalized layer thickness from two equipment.} Boxplot graphs of normalized metrology/inline measurements from both equipment types considered in the analysis.}\label{fig: boxplotVM2}
\end{figure}

The data used in this work is collected from two different etching equipment types from the same vendor. The data set is restricted to a specific etching recipe that was transferred from one equipment to the other and now runs regularly on both equipment. Raw sensor measurements in form of time series data and their corresponding metrology/inline measurements over a period of 3 years are considered. 
\begin{itemize}
    \item The equipment type 1 with 35 activated sensors - older equipment type hence higher number of samples ($\sim$10 000) and original tool to run the specific recipe- is selected as source;
    \item the equipment type 2 with 55 activated sensors - newer equipment type with $\sim$6000 data samples - is defined as target. 
\end{itemize}
The following preprocessing steps were applied to the collected time series sensor data, with each equipment treated separately due to its heterogeneous nature:
\begin{enumerate}
    \item removal of constant features;
     \item removal of features that show small fluctuation that can be detect as noise (variations smaller than $0.01$) and a constant behavior underneath the noise;
      \item removal of samples showing label outliers based on interquantile range;
      \item removal of samples where the length of the time series lies below or above 25 percent respective 75 percent quantile of time series length;
    \item equal-distributed upsampling of timestamps and feature values to generate time series with equal length.
\end{enumerate}
33 features for equipment type 1 respective 49 for equipment type 2 are finally selected as input features. No significant label shift is detected, see Figure \ref{fig: boxplotVM2}.

\subsection{Experimental Design}
Virtual metrology is modeled as prediction task with sensor data as input mapped to a single continuous metrology value. Due to the heterogeneous nature of the data representation from equipment type 1 and 2, no common model can be used without additional transfer. 
We present the analysis in the following order:
\begin{enumerate}
    \item DBACS is trained and tested as domain adaptation model using autoencoder to align the original input features to enable usage of a dedicated pretrained model;
    \item PCA and CCA are selected as benchmark models; for the heterogeneous VM task the alignment happens by creating a common (latent) feature space that is then used to train a common model. CORAL on the latent features is tested as final combination for both PCA and CCA.
\end{enumerate}
For training, distribution comparison and alignment evaluation the following metrics are considered:
\begin{itemize}
    \item Mean absolute error (MAE) is used as performance based loss; Adam optimizer is applied for training;
    \item To test inner versus outer domain distance - the divergence between data selected from source and data selected from target domain - we use Frechet inception distance (FID);
    \item 5-fold cross validation is applied, hence split both data sets into 5 subsets each and using 4 merged sets as train and 1 as test set per fold. Architectures of all models stay fixed for all 5 folds;
    \item pearson correlation of features is tested after alignment.
 \end{itemize}   
For the correlation analysis we use the function implementation available in python module \emph{numpy} \cite{numpy} and for PCA and CCA we use existing function implementation in the python module \emph{scikit-learn} \cite{scikit-learn}. For CORAL we use the implementations from the python module \emph{transfertools} \cite{transfertools}. DBACS is trained using the described adversarial training approach. PCA and CCA expect a two dimensional input, hence we keep the original data and reshape the 3 dimensional sample into a two dimensional one by treating each value at each time step as separate sample. The selected number of latent features are based on the variation coverage of both domains. In the following, model details and hyperparameters choices are reported.
\paragraph{DBCAS} 1DCNN is chosen since it is simple but proven to be well performing for time series data \cite{gentner2021}. 
\begin{itemize}
    \item The predictor consists of 3 convolutional layers (dimension 32, 16, 8 and kernel size 53, 33 and 33), followed by one max pooling layer, a flattening layer and two dense layers (dimension 16 and 1, Leaky ReLU activation except sigmoid output). 
    \item The domain discriminators both have the same architecture besides the respective input shape: 3 convolutional layers (dimension 24, 16 and 8, kernel size 17), causal padding and leaky ReLU activation function, max pooling of size 4 and 2 times 2, followed by a flattening layer and 6 dense layers (dimension 512, 256, 128, 64, 32, 1, Leaky ReLU activation and linear output).  \item Both aligners consist of 6 convolutional layers, first 5 followed by Leaky ReLU activation function, the final output is kept linear. The aligner that maps target domain to source domain has filter size 48, 42, 36, 32, 32 and final filter size is set to number of features of the source domain; kernel size is 37, 37, 37, 37, 57 and 7. Upsampling with size 3 and 2 is done after 4th and 5th layer block. The aligner that maps source domain to target domain has filter size 32, 36, 42, 46, 48 and final filter size is set to number of features of the target domain; kernel size is also 37, 37, 37, 37, 57 and 7. Upsampling with size 3 and 2 is also done after 4th and 5th layer block. 
\end{itemize}    
For an improved initialization both aligners are pretrained separately using SSIM loss \cite{ssim}: therefore, we select sample pairs from source and target based on closest label value.
\paragraph{PCA and CORAL} The 3 dimensional input is reshaped into a two dimensional one by treating each value at each time step as separate sample. For each equipment type, we select the first 10 principal components in order to cover around $95\%$ variance and to create same dimensional input space. For equipment type 1 we cover $97\%$ of the variance and for equipment type 2 we cover $94\%$. 1DCNN prediction model with reduced number of features based on PCA (reshaped back to 3 dimensions) of source and target domain is used as prediction model.

\paragraph{CCA} 27 canonical components (CCs) are kept since it shows the most stable results in our experiments. The original data is reshaped from 3 dimensional input into a two dimensional one by treating each value at each time step as separate sample. 1DCNN prediction model with reduced number of features based on CCA (reshaped back to 3 dimensions) of source and target domain is used as prediction model.

\section{Experimental Results} \label{section:results}
\begin{figure}[htb]
\centering
\includegraphics[width=0.33\textwidth]{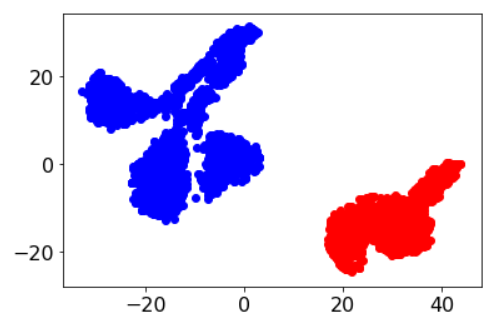}
\includegraphics[width=0.33\textwidth]{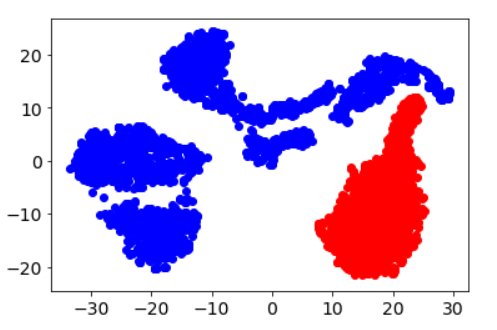}
\includegraphics[width=0.33\textwidth]{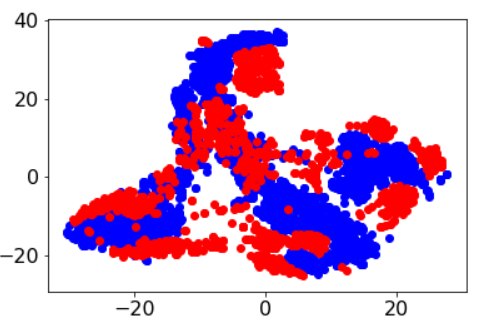}
\caption{\textbf{T-SNE visualization before and after alignment with DBACS.} Graphical t-SNE representation of source and target domain in different stages of the alignment process: (a) shows features mapped by a randomly initialized aligner, (b) after the pretraining of the aligner and (c) after DA with DBACS is done. The source is colored in blue and contains data from equipment type 1, the target is colored red and contains data from the equipment type 2. The axes are dimensionless. The effect of the adaptation of the input features after DBACS is applied during training. The adaptation brings the distributions of the target domain closer and finally target overlaps source domain.}\label{fig: tsneVM2}
\end{figure}

\begin{figure}[htb]
\centering
\includegraphics[width=0.4\textwidth]{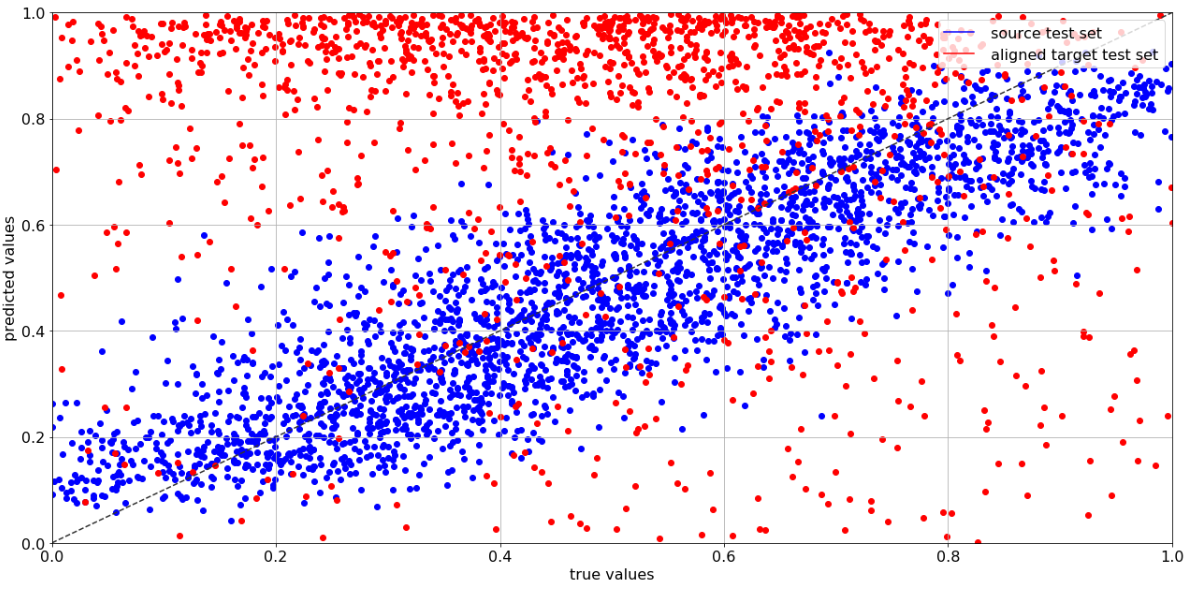} 
\includegraphics[width=0.4\textwidth]{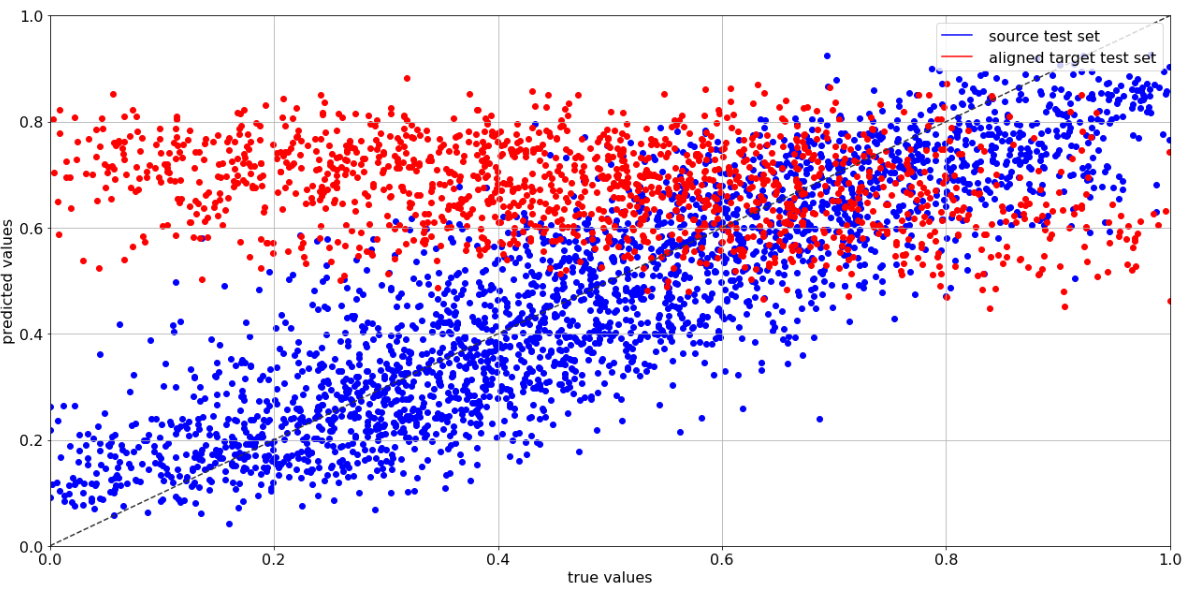}  
\includegraphics[width=0.7\textwidth]{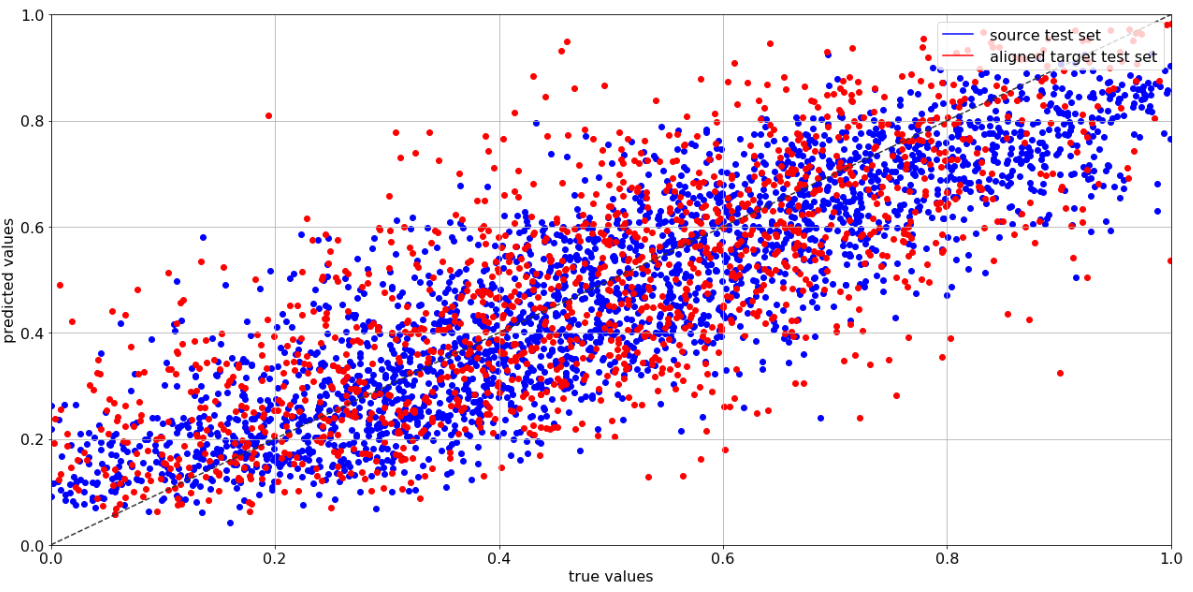}
\caption{True versus predicted scatter plot for DBACS before and after alignment. The graph shows predictions of aligned target data after mapped to source space by a randomly initialized aligner, after the pretraining of the aligner and predictions of aligned target data after DA training with DBACS is done. Only test data is presented, the test source data is colored in blue, the the aligned target test data is colored in red.}\label{fig: pred_vm2}
\end{figure}

\begin{figure}[ht]
\centering
\includegraphics[width=1\textwidth]{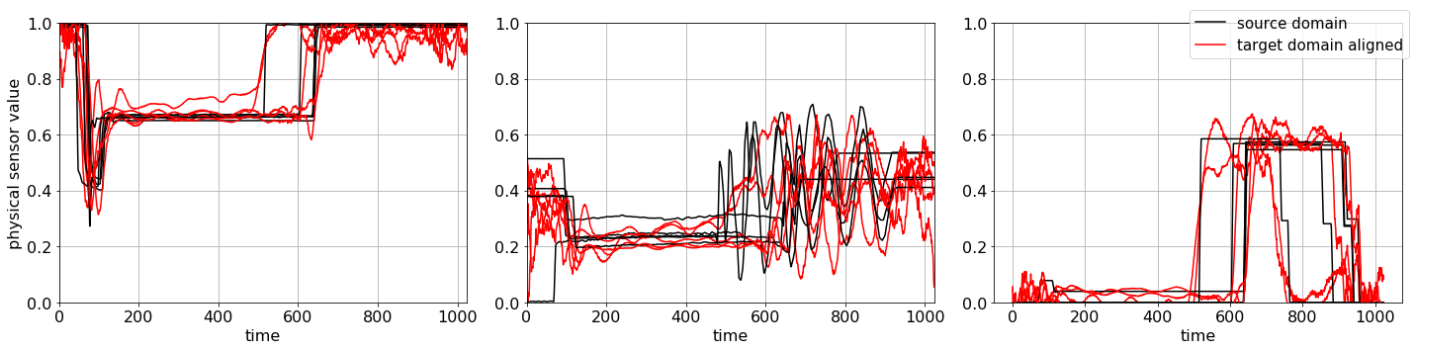}
\includegraphics[width=1\textwidth]{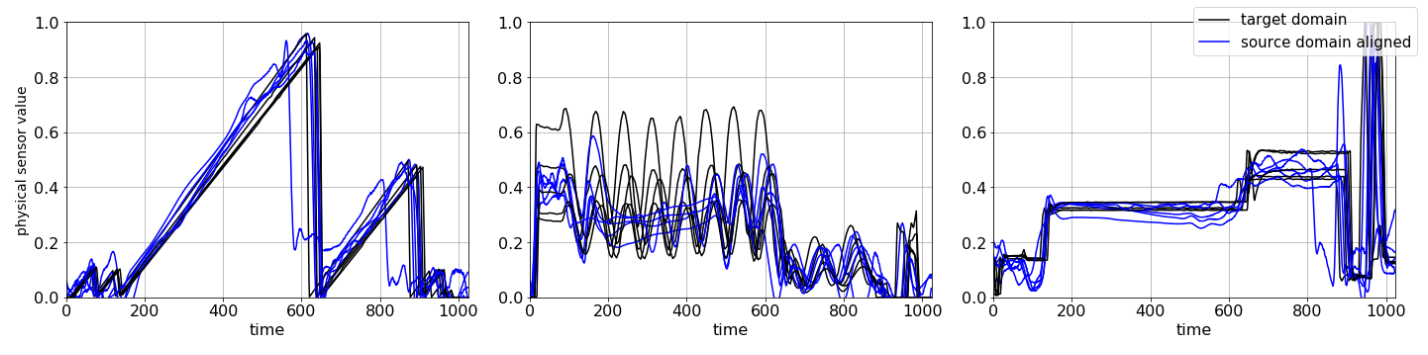}
\caption{Aligner $F$ and $G$ visualizations of 2 times 3 raw sensor measurements of both equipment types before and after the corresponding alignment. The graph shows results for trained aligner $F$ and mapped sensor signals from target to source domain in red and compares it to corresponding original source sensor signals plotted in black. It also shows results from trained aligner $G$ and mapped sensor signals from source to target domain in blue and compares it to corresponding original target sensor signals plotted in black. A good alignment is visible as well. The x axis shows the timestamps of the sensor signals, y axis the sensor measurement values.}\label{fig: sensor_vm2}
\end{figure}

Table \ref{tab: results_vm2} shows the average 5 fold CV results for DBACS compared to the dedicated lower bound values meaning performance errors for dedicated models trained only on source and only on target data.
\begin{table}[ht]
\begin{center}
\begin{tabular}{p{2cm}|p{2cm}|p{2cm}||p{2cm}|p{2cm}}
\multicolumn{5}{c}{\textbf{DBACS MAE for source and aligned target}} \\
 & \multicolumn{2}{c||}{Source domain} & \multicolumn{2}{c}{Target domain} \\
\hline
 & Train &Test& Train & Test \\ \hline \hline
Lower Bound & $0.084$ &	$0.094$	& $0.102$ & $0.128$	 \\  \hline
DBACS & $0.084$ &	$0.094$	& \textbf{$0.102$} &	\textbf{$0.131$} \\ 
\end{tabular}
\end{center}
\caption{\textbf{DBACS performance errors for source and aligned target.} Source and aligned target data DBACS training and test scores average over 5 fold CV. Target data is mapped to source domain using trained aligner $F$ from DBACS and evaluated after the mapping using the VM prediction model trained on source. Lower bound prediction models are dedicated meaning trained only on source train data and evaluated only on source test data respective trained only on target train data and evaluated only on target test data.}\label{tab: results_vm2}
\end{table}
The numbers given in Table \ref{tab: results_vm2} confirm the visual convergence seen in the t-SNE plot in Figure \ref{fig: tsneVM2}. This is supported by frechet inception distance (FID) $0.01$ for outer domain distance after alignment compared to FID inner domain distance close to 0 for equipment type 1 as well as for equipment type 2.
Next, Figure \ref{fig: pred_vm2} shows true versus predicted values of different alignment states - randomly initialized aligner, after the pretraining of the aligner and after DA training with DBACS).  Again, the visualization supports the results presented in Table \ref{tab: results_vm2}: Enabeling usage of a dedicated source model to mapped target data for high accuracy predictions.
A visualization of both aligners output is presented in Figure \ref{fig: sensor_vm2} and compared to original domain sensor signals.\\

Next, we present results for PCA analysis in Table \ref{tab: results_pca_vm2}. Optional DA with CORAL on top shows slightly improved results if model is trained on data from both domains. The FID score for outer domain distance after PCA + CORAL on the latent features generated by PCA is significant lower than before with $0.0001$ for train and $0.001$ for test. Only the first two principal components show a correlation higher than $r=0.5$.\\
\begin{table}[ht]
\begin{center}
\begin{tabular}{c|c|c||c|c}
\multicolumn{5}{c}{\textbf{VM prediction model performance for PCA based principle components}} \\
 & \multicolumn{2}{c||}{Source domain} & \multicolumn{2}{c}{Target domain} \\
\hline
 & Train MAE &Test MAE& Train MAE & Test MAE \\ \hline \hline
PCA(source) & $0.09$&	$0.09$&	$0.32$&	$0.33$\\ \hline
PCA(target) & $0.47$&	$0.47$&	$0.12$&	$0.13$\\ \hline
PCA(both) & $0.10$	& $0.14$ & 	$0.09$	& $0.14$\\ \hline
PCA+CORAL(both) &  $0.08$	& $0.09$ & 	$0.12$	& $0.13$\\
\end{tabular}
\end{center}
\caption{\textbf{VM prediction model performance for PCA based principle components.} Results for VM prediction models that are trained with reduced number of latent features that are created via PCA.}\label{tab: results_pca_vm2}
\end{table}

For CCA, the performance is presented in Table \ref{tab: results_cca_vm2}. Optional DA with CORAL on top shows improved results since model training for both domains is enabled and can be executed using CCs. The FID score for outer domain distance after CCA + CORAL on the latent features generated by CCA is again significant lower than before with very close to $0$ for train and test. The first five CCs have a correlation higher than $r=0.5$.\\
\begin{table}[ht]
\begin{center}
\begin{tabular}{c|c|c||c|c}
\multicolumn{5}{c}{\textbf{VM prediction model performance for CCA based canonical components}} \\
 & \multicolumn{2}{c||}{Source domain} & \multicolumn{2}{c}{Target domain} \\
\hline
 & Train MAE &Test MAE& Train MAE & Test MAE \\ \hline \hline
CCA(source) & $0.12$&	$0.13$&	$0.29$&	$0.29$\\ \hline
CCA(target) & $0.29$&	$0.29$&	$0.10$&	$0.14$\\ \hline
CCA(both) & $0.10$	& $0.13$ & 	$0.12$	& $0.14$\\\hline
CCA+CORAL(both) & $0.07$	& $0.08$ & 	$0.13$	& $0.13$\\
\end{tabular}
\end{center}
\caption{\textbf{VM prediction model performance for CCA based canonical components.} Results for VM prediction models that are trained with latent features created via CCA.}\label{tab: results_cca_vm2}
\end{table}

\section{Equipment Matching Experiments}\label{sec:discussion}
Having with DBACS a methodology that allows parallel training and transfer in both directions - source to target but also target to source - mis- or abnormal behavior detected for aligned data can be compared to normal as well as abnormal data from source. These kind of comparisons enables equipment matching for nonidentical equipment with heterogeneous data representations.

\begin{figure}[ht]
\centering
\includegraphics[width=1\textwidth]{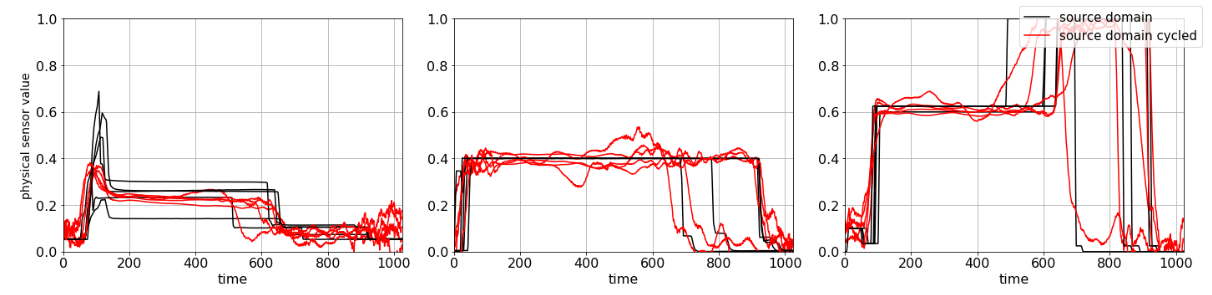}
\includegraphics[width=1\textwidth]{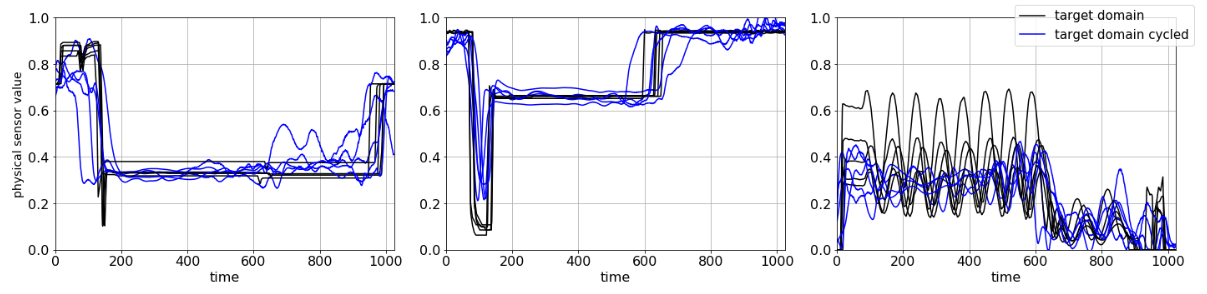}
\caption{\textbf{Aligner $F$ and $G$ visualizations of 2 times 3 cycled raw sensor measurements of both equipment types in its original form as well as after its bijective mapping.} The first graph shows results for source signals and cycled source signals from source to target to source domain. The cycled signals are plotted in red and compared to its original source sensor signals plotted in black. The second graph shows results for target signals and cycled target signals from target to source back to target domain. The cycled signals are plotted in blue and compared to its original target sensor signals plotted in black. The x axis shows the timestamps of the sensor signals, y axis the sensor measurement values.}\label{fig: signals_cycled}
\end{figure}

First, we compare source signals with its cycled signals on the signal shape itself as well target signals with its cycled target signals. Examples from both are presented in Figure \ref{fig: signals_cycled}.\\

\begin{figure}[ht]
\centering
\includegraphics[width=1\textwidth]{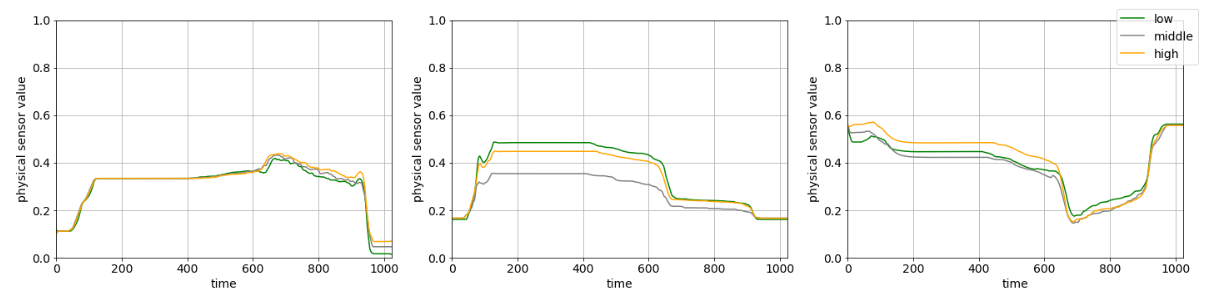}
\caption{\textbf{Comparison of raw source sensor measurements via barycenter average grouped into low, middle high label values. }Graphical representation of euclidean barycenter averages for 3 example sensors of the source domain. The x axis shows the timestamps of the sensor signals, y axis the sensor measurement values. Example sensor measurements of samples corresponding to low label values - meaning values smaller $0.1$ - are plotted in green, example sensor measurements of samples corresponding to middle therefore preferred label values - meaning values around $0.5$ - are plotted in grey and example sensor measurements of samples corresponding to high label values - meaning values higher than $0.9$ - are plotted in orange. Sensor offsets for deviating metrology measurements are clearly visible.}\label{fig: source_low_middle_high}
\end{figure}

Next, we check differences within source domain of samples having a high, middle and low prediction value. This helps to better understand univariate feature behavior for source. The middle prediction is the preferred and targeted one. Figure \ref{fig: source_low_middle_high} shows euclidean barycenter averages of tree example signals from source domain for low, middle and high label values. Sensor offsets for deviating metrology measurements are clearly visible for some of the signals.\\

\begin{figure}[ht]
\centering
\includegraphics[width=1\textwidth]{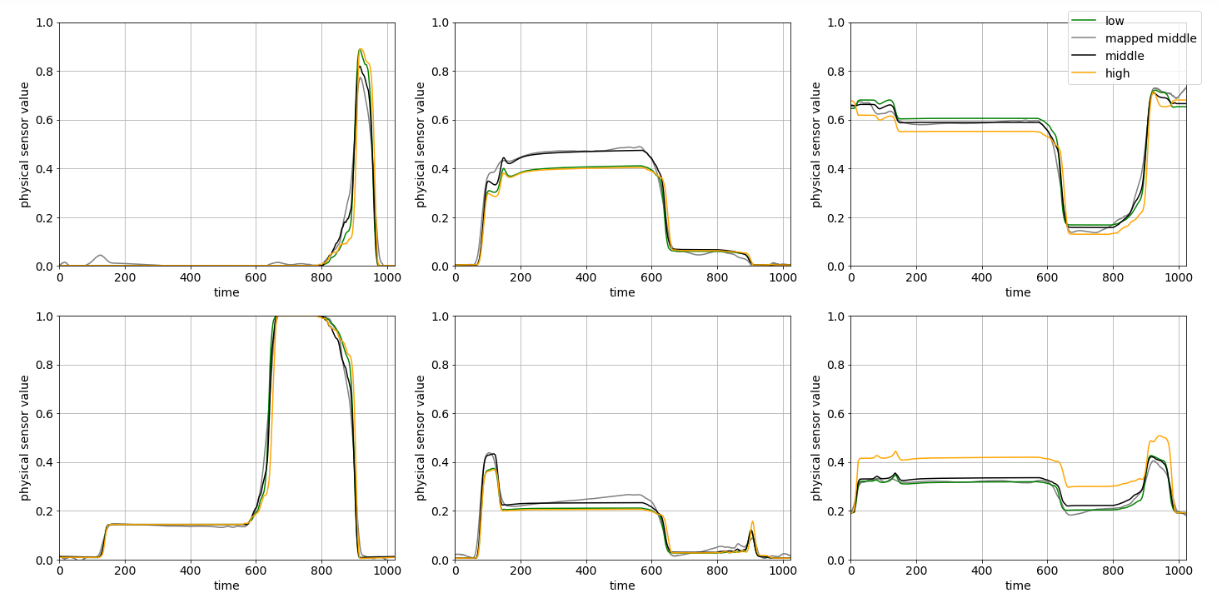}
\caption{\textbf{Visualization of equipment matching: mapped source sensors vs. target sensors grouped into low, middle high label values.} Graphical representation of 6 sensor signals. The plots show euclidean barycenter averages of signals. Groups are defined by their metrology values and plotted in different colors.  The x axis shows the timestamps of the sensor signals, y axis the sensor measurement values. Selected target sensor measurements of samples corresponding to low label values - meaning values smaller $0.1$ - are plotted in green, example target sensor measurements of samples corresponding to middle therefore preferred label values - meaning values around $0.5$ - are plotted in black and example sensor measurements of samples corresponding to high label values - meaning values higher than $0.9$ - are plotted in orange. Mapped sensor signals corresponding to source samples with middle label values are colored black.  Sensor offsets for deviating metrology measurements are clearly visible from middle target as well as mapped middle source signals.}\label{fig: equip_match}
\end{figure}

For final equipment matching, we compare preferred shape of signals from the source domain meaning signals with metrology measurements close to target $0.5$ (see Figure \ref{fig: source_low_middle_high} to corresponding as well as deviating signals from target domain. Therefore, we use the DBACS to map selected source signals into the target domain. Different sensors measurements and their euclidean barycenter averages of groups according to low, middle and high metrology measurements are shown in Figure \ref{fig: equip_match} and compared to mapped sensor signals (source to target) corresponding to the middle meaning preferred metrology group in the source domain.

\section{Conclusion and Future Work}\label{sec:conclusions}
The paper presents DBACS, a Deep Learning approach that is able to deal with heterogeneous domain adaptation while allowing comparison of aligned signals for a VM use case in semiconductor manufacturing. Linear transformation methods from subspace alignment and multi-view learning are selected as benchmarks and show comparable results when training with data from both domains is possible. Especially for classification tasks, the correlation within CCA can be further exploited for cross-modal or mate-based retrieval. A big advantage of DBACS is the presented combination of domain adaptation with matching, two of the main approaches for standardization and scalability in the semiconductor field. 

Envisioned future work could go in the direction of root cause analysis based on the matching results. Another important step could to enrich the data with more equipment for multi-source or multi-target alignment. Other applications from semiconductor manufacturing like predictive maintenance and defect classification could be involved and tested for example against computer vision inspired state-of-the-art transfer learning benchmark models like pseudo-labeling. Since only offline model training is executed (training time is not a critical aspect of VM here), online model training could also be explored in that context.

\section*{Acknowledgment}
Infineon Technologies AG is gratefully acknowledged for the financial support of this research. 
The Italian Government PNRR iniatiatives 'Partenariato 11: Made in Italy circolare e sostenibile' and 'Ecosistema dell'Innovazione  - iNest' are also gratefully acknowledged for partially financing this research activity.

\bibliographystyle{abbrvnat}
\bibliography{preprint}  

\begin{thebibliography}{54}
\providecommand{\natexlab}[1]{#1}
\providecommand{\url}[1]{\texttt{#1}}
\expandafter\ifx\csname urlstyle\endcsname\relax
  \providecommand{\doi}[1]{doi: #1}\else
  \providecommand{\doi}{doi: \begingroup \urlstyle{rm}\Url}\fi

\bibitem[Alipour and Tahmoresnezhad(2022)]{Alipour2022HeterogeneousDA}
N.~Alipour and J.~Tahmoresnezhad.
\newblock Heterogeneous domain adaptation with statistical distribution
  alignment and progressive pseudo label selection.
\newblock \emph{Appl. Intell.}, 52:\penalty0 8038--8055, 2022.
\newblock \doi{https://doi.org/10.1007/s10489-021-02756-x}.

\bibitem[{Azamfar} et~al.(2020){Azamfar}, {Li}, and {Lee}]{azamfar2020}
M.~{Azamfar}, X.~{Li}, and J.~{Lee}.
\newblock Deep learning-based domain adaptation method for fault diagnosis in
  semiconductor manufacturing.
\newblock \emph{IEEE Transactions on Semiconductor Manufacturing}, 33\penalty0
  (3):\penalty0 445--453, 2020.
\newblock \doi{https://doi.org/10.1109/TSM.2020.2995548}.

\bibitem[Chapman and Wang(2021)]{Chapman2021}
J.~Chapman and H.-T. Wang.
\newblock Cca-zoo: A collection of regularized, deep learning based, kernel,
  and probabilistic cca methods in a scikit-learn style framework.
\newblock \emph{Journal of Open Source Software}, 6\penalty0 (68):\penalty0
  3823, 2021.
\newblock \doi{https://doi.org/10.21105/joss.03823}.

\bibitem[Chen et~al.(2020)Chen, Zhao, Pang, and Lin]{CHEN2020192}
C.-H. Chen, W.-D. Zhao, T.~Pang, and Y.-Z. Lin.
\newblock Virtual metrology of semiconductor pvd process based on combination
  of tree-based ensemble model.
\newblock \emph{ISA Transactions}, 103:\penalty0 192 -- 202, 2020.
\newblock \doi{https://doi.org/10.1016/j.isatra.2020.03.031}.

\bibitem[Chen et~al.(2005)Chen, Wu, Lin, Ko, Lo, Wang, Yu, and
  Liang]{chen2005vm}
P.~Chen, S.~Wu, J.~Lin, F.~Ko, H.~Lo, J.~Wang, C.~Yu, and M.~Liang.
\newblock Virtual metrology: a solution for wafer to wafer advanced process
  control.
\newblock In \emph{ISSM 2005, IEEE International Symposium on Semiconductor
  Manufacturing, 2005.}, pages 155--157, 2005.
\newblock \doi{https://doi.org/10.1109/ISSM.2005.1513322}.

\bibitem[Chen et~al.(2016)Chen, Zhang, Ding, Shardt, and Hu]{CHEN2016cca}
Z.~Chen, K.~Zhang, S.~X. Ding, Y.~A. Shardt, and Z.~Hu.
\newblock Improved canonical correlation analysis-based fault detection methods
  for industrial processes.
\newblock \emph{Journal of Process Control}, 41:\penalty0 26--34, 2016.
\newblock \doi{https://doi.org/10.1016/j.jprocont.2016.02.006}.

\bibitem[Chien et~al.(2022{\natexlab{a}})Chien, Hung, and Liao]{9749610}
C.-F. Chien, W.-T. Hung, and E.~T.-Y. Liao.
\newblock Redefining monitoring rules for intelligent fault detection and
  classification via cnn transfer learning for smart manufacturing.
\newblock \emph{IEEE Transactions on Semiconductor Manufacturing}, 35\penalty0
  (2):\penalty0 158--165, 2022{\natexlab{a}}.
\newblock \doi{https://doi.org/10.1109/TSM.2022.3164904}.

\bibitem[Chien et~al.(2022{\natexlab{b}})Chien, Hung, Pan, and
  Van~Nguyen]{chien2022decision}
C.-F. Chien, W.-T. Hung, C.-W. Pan, and T.~H. Van~Nguyen.
\newblock Decision-based virtual metrology for advanced process control to
  empower smart production and an empirical study for semiconductor
  manufacturing.
\newblock \emph{Computers \& Industrial Engineering}, page 108245,
  2022{\natexlab{b}}.

\bibitem[Chouichi et~al.(2020)Chouichi, Blue, Yugma, and
  Pasqualini]{chouichi2020chamber}
A.~Chouichi, J.~Blue, C.~Yugma, and F.~Pasqualini.
\newblock Chamber-to-chamber discrepancy detection in semiconductor
  manufacturing.
\newblock \emph{IEEE Transactions on Semiconductor Manufacturing}, 33\penalty0
  (1):\penalty0 86--95, 2020.
\newblock \doi{https://doi.org/10.1109/TSM.2020.2965288}.

\bibitem[Courty et~al.(2017)Courty, Flamary, Tuia, and
  Rakotomamonjy]{courty2016optimal}
N.~Courty, R.~Flamary, D.~Tuia, and A.~Rakotomamonjy.
\newblock Optimal transport for domain adaptation.
\newblock \emph{IEEE Transactions on Pattern Analysis and Machine
  Intelligence}, 39\penalty0 (9):\penalty0 1853--1865, 2017.
\newblock \doi{https://doi.org/10.1109/TPAMI.2016.2615921}.

\bibitem[Fan et~al.(2020)Fan, Hsu, Tsai, He, and Cheng]{9066890}
S.-K.~S. Fan, C.-Y. Hsu, D.-M. Tsai, F.~He, and C.-C. Cheng.
\newblock Data-driven approach for fault detection and diagnostic in
  semiconductor manufacturing.
\newblock \emph{IEEE Transactions on Automation Science and Engineering},
  17\penalty0 (4):\penalty0 1925--1936, 2020.
\newblock \doi{https://doi.org/10.1109/TASE.2020.2983061}.

\bibitem[Fang et~al.(2022)Fang, Lu, Liu, and Zhang]{fang2022hetero}
Z.~Fang, J.~Lu, F.~Liu, and G.~Zhang.
\newblock Semi-supervised heterogeneous domain adaptation: Theory and
  algorithms.
\newblock \emph{IEEE Transactions on Pattern Analysis and Machine
  Intelligence}, 45\penalty0 (1):\penalty0 1087--1105, 2022.
\newblock \doi{https://doi.org/10.1109/TPAMI.2022.3146234}.

\bibitem[Farahani et~al.(2020)Farahani, Fatehi, Nadali, and
  Shoorehdeli]{farahani2020novel}
H.~S. Farahani, A.~Fatehi, A.~Nadali, and M.~A. Shoorehdeli.
\newblock A novel method for designing transferable soft sensors and its
  application.
\newblock \emph{arXiv}, 2020.
\newblock \doi{https://doi.org/10.48550/arxiv.2008.02186}.

\bibitem[Feng et~al.(2019)Feng, Jia, Zhu, Moyne, Iskandar, and
  Lee]{feng2019online}
J.~Feng, X.~Jia, F.~Zhu, J.~Moyne, J.~Iskandar, and J.~Lee.
\newblock An online virtual metrology model with sample selection for the
  tracking of dynamic manufacturing processes with slow drift.
\newblock \emph{IEEE Transactions on Semiconductor Manufacturing}, 32\penalty0
  (4):\penalty0 574--582, 2019.
\newblock \doi{https://doi.org/10.1109/TSM.2019.2942768}.

\bibitem[Fernando et~al.(2013)Fernando, Habrard, Sebban, and
  Tuytelaars]{basura2013sa}
B.~Fernando, A.~Habrard, M.~Sebban, and T.~Tuytelaars.
\newblock Unsupervised visual domain adaptation using subspace alignment.
\newblock In \emph{2013 IEEE International Conference on Computer Vision},
  pages 2960--2967, 2013.
\newblock \doi{https://doi.org/10.1109/ICCV.2013.368}.

\bibitem[Ganin et~al.(2016)Ganin, Ustinova, Ajakan, Germain, Larochelle,
  Laviolette, Marchand, and Lempitsky]{ganin2015domainadversarial}
Y.~Ganin, E.~Ustinova, H.~Ajakan, P.~Germain, H.~Larochelle, F.~Laviolette,
  M.~Marchand, and V.~Lempitsky.
\newblock Domain-adversarial training of neural networks.
\newblock \emph{J. Mach. Learn. Res.}, 17\penalty0 (1):\penalty0 2096–2030,
  Jan. 2016.
\newblock \doi{https://doi.org/10.5555/2946645.2946704}.

\bibitem[Gentner et~al.(2020)Gentner, Carletti, Kyek, Susto, and
  Yang]{gentner2020}
N.~Gentner, M.~Carletti, A.~Kyek, G.~A. Susto, and Y.~Yang.
\newblock Enhancing scalability of virtual metrology: A deep learning-based
  approach for domain adaptation.
\newblock In \emph{Proceedings of the 2020 Winter Simulation Conference}, pages
  1898--1909, 2020.
\newblock \doi{https://doi.org/10.1109/WSC48552.2020.9383945}.

\bibitem[Gentner et~al.(2021)Gentner, Carletti, Kyek, Susto, and
  Yang]{gentner2021}
N.~Gentner, M.~Carletti, A.~Kyek, G.~A. Susto, and Y.~Yang.
\newblock Dbam: Making virtual metrology/soft sensing with time series data
  scalable through deep learning.
\newblock \emph{Control Engineering Practice}, 116:\penalty0 104914, 2021.
\newblock \doi{https://doi.org/10.1016/j.conengprac.2021.104914}.

\bibitem[Gulrajani et~al.(2017)Gulrajani, Ahmed, Arjovsky, Dumoulin, and
  Courville]{gulrajani2017ganimproved}
I.~Gulrajani, F.~Ahmed, M.~Arjovsky, V.~Dumoulin, and A.~Courville.
\newblock Improved training of wasserstein gans.
\newblock In \emph{Proceedings of the 31st International Conference on Neural
  Information Processing Systems}, NIPS’17, page 5769–5779, Red Hook, NY,
  USA, 2017. Curran Associates Inc.
\newblock \doi{https://doi.org/10.48550/arxiv.1704.00028}.

\bibitem[Hardoon et~al.(2004)Hardoon, Szedmak, and Shawe-Taylor]{cca2004}
D.~R. Hardoon, S.~Szedmak, and J.~Shawe-Taylor.
\newblock Canonical correlation analysis: An overview with application to
  learning methods.
\newblock \emph{Neural Computation}, 16\penalty0 (12):\penalty0 2639--2664,
  2004.
\newblock \doi{https://doi.org/10.1162/0899766042321814}.

\bibitem[Harris et~al.(2020)Harris, Millman, van~der Walt, Gommers, Virtanen,
  Cournapeau, Wieser, Taylor, Berg, Smith, Kern, Picus, Hoyer, van Kerkwijk,
  Brett, Haldane, del R{\'{i}}o, Wiebe, Peterson, G{\'{e}}rard-Marchant,
  Sheppard, Reddy, Weckesser, Abbasi, Gohlke, and Oliphant]{numpy}
C.~R. Harris, K.~J. Millman, S.~J. van~der Walt, R.~Gommers, P.~Virtanen,
  D.~Cournapeau, E.~Wieser, J.~Taylor, S.~Berg, N.~J. Smith, R.~Kern, M.~Picus,
  S.~Hoyer, M.~H. van Kerkwijk, M.~Brett, A.~Haldane, J.~F. del R{\'{i}}o,
  M.~Wiebe, P.~Peterson, P.~G{\'{e}}rard-Marchant, K.~Sheppard, T.~Reddy,
  W.~Weckesser, H.~Abbasi, C.~Gohlke, and T.~E. Oliphant.
\newblock Array programming with {NumPy}.
\newblock \emph{Nature}, 585\penalty0 (7825):\penalty0 357--362, Sept. 2020.
\newblock \doi{https://doi.org/10.1038/s41586-020-2649-2}.

\bibitem[Heng et~al.(2021)Heng, Liao, Didari, and Rajagopal]{applied2020}
H.~Heng, T.~Liao, S.~Didari, and H.~Rajagopal.
\newblock \emph{Chamber matching with neural networks in semiconductor
  equipment tools}.
\newblock Applied Materials, Inc., US Patent 11,133,204, 2021.

\bibitem[Hilleringmann(1996)]{hilleringmann}
U.~Hilleringmann.
\newblock \emph{Silizium-Halbleitertechnologie}.
\newblock Springer, 1996.

\bibitem[Jolliffe(2010)]{pca}
I.~Jolliffe.
\newblock \emph{Principal Component Analysis}.
\newblock Springer Series in Statistics. Springer New York, 2010.
\newblock ISBN 9781441929990.

\bibitem[Kang(2017)]{kang2017effectiveness}
S.~Kang.
\newblock On effectiveness of transfer learning approach for neural
  network-based virtual metrology modeling.
\newblock \emph{IEEE Transactions on Semiconductor Manufacturing}, 31\penalty0
  (1):\penalty0 149--155, 2017.
\newblock \doi{https://doi.org/10.1109/TSM.2017.2787550}.

\bibitem[Kang et~al.(2016)Kang, Kim, and Cho]{kang2016vmfeature}
S.~Kang, D.~Kim, and S.~Cho.
\newblock Efficient feature selection-based on random forward search for
  virtual metrology modeling.
\newblock \emph{IEEE Transactions on Semiconductor Manufacturing}, 29\penalty0
  (4):\penalty0 391--398, 2016.
\newblock \doi{https://doi.org/10.1109/TSM.2016.2594033}.

\bibitem[Lee and Kim(2020)]{lee2020recurrent}
K.~B. Lee and C.~O. Kim.
\newblock Recurrent feature-incorporated convolutional neural network for
  virtual metrology of the chemical mechanical planarization process.
\newblock \emph{Journal of Intelligent Manufacturing}, 31\penalty0
  (1):\penalty0 73--86, 2020.
\newblock \doi{https://doi.org/10.1007/s10845-018-1437-4}.

\bibitem[Li et~al.(2020)Li, Wang, Zhang, Li, Darrell, Keutzer, and
  Zhao]{li2020learning}
B.~Li, Y.~Wang, S.~Zhang, D.~Li, T.~Darrell, K.~Keutzer, and H.~Zhao.
\newblock Learning invariant representations and risks for semi-supervised
  domain adaptation.
\newblock \emph{arXiv}, 2020.
\newblock \doi{https://doi.org/10.48550/arxiv.2010.04647}.

\bibitem[Lu et~al.(2019)Lu, Liu, and Hsu]{lu2019pix}
Y.-W. Lu, K.-L. Liu, and C.-Y. Hsu.
\newblock Conditional generative adversarial network for defect classification
  with class imbalance.
\newblock In \emph{2019 IEEE International Conference on Smart Manufacturing,
  Industrial \& Logistics Engineering (SMILE)}, pages 146--149, 2019.
\newblock \doi{https://doi.org/10.1109/SMILE45626.2019.8965320}.

\bibitem[Lynn et~al.(2009)Lynn, Ringwood, Ragnoli, McLoone, and
  MacGearailty]{lynn2009virtual}
S.~Lynn, J.~Ringwood, E.~Ragnoli, S.~McLoone, and N.~MacGearailty.
\newblock Virtual metrology for plasma etch using tool variables.
\newblock In \emph{2009 IEEE/SEMI Advanced Semiconductor Manufacturing
  Conference}, pages 143--148. IEEE, 2009.
\newblock \doi{https://doi.org/10.1109/ASMC.2009.5155972}.

\bibitem[Maggipinto et~al.(2018)Maggipinto, Masiero, Beghi, and
  Susto]{MAGGIPINTO2018}
M.~Maggipinto, C.~Masiero, A.~Beghi, and G.~A. Susto.
\newblock A convolutional autoencoder approach for feature extraction in
  virtual metrology.
\newblock \emph{Procedia Manufacturing}, 17:\penalty0 126--133, 2018.
\newblock \doi{https://doi.org/10.1016/j.promfg.2018.10.023}.
\newblock 28th International Conference on Flexible Automation and Intelligent
  Manufacturing (FAIM2018), June 11-14, 2018, Columbus, OH, USAGlobal
  Integration of Intelligent Manufacturing and Smart Industry for Good of
  Humanity.

\bibitem[Maggipinto et~al.(2019)Maggipinto, Beghi, McLoone, and
  Susto]{maggipinto2019deepvm}
M.~Maggipinto, A.~Beghi, S.~McLoone, and G.~A. Susto.
\newblock Deepvm: A deep learning-based approach with automatic feature
  extraction for 2d input data virtual metrology.
\newblock \emph{Journal of Process Control}, 84:\penalty0 24--34, 2019.
\newblock \doi{https://doi.org/10.1016/j.jprocont.2019.08.006}.

\bibitem[May and Spanos(2006)]{spanos2006}
G.~May and C.~Spanos.
\newblock \emph{Fundamentals of Semiconductor Manufacturing and Process
  Control}.
\newblock IEEE Press. John Wiley \& Sons, 2006.

\bibitem[Park and Kim(2016)]{park2016virtual}
C.~Park and S.~B. Kim.
\newblock Virtual metrology modeling of time-dependent spectroscopic signals by
  a fused lasso algorithm.
\newblock \emph{Journal of Process Control}, 42:\penalty0 51--58, 2016.
\newblock \doi{https://doi.org/10.1016/j.jprocont.2016.04.002}.

\bibitem[Pedregosa et~al.(2011)Pedregosa, Varoquaux, Gramfort, Michel, Thirion,
  Grisel, Blondel, Prettenhofer, Weiss, Dubourg, Vanderplas, Passos,
  Cournapeau, Brucher, Perrot, and Duchesnay]{scikit-learn}
F.~Pedregosa, G.~Varoquaux, A.~Gramfort, V.~Michel, B.~Thirion, O.~Grisel,
  M.~Blondel, P.~Prettenhofer, R.~Weiss, V.~Dubourg, J.~Vanderplas, A.~Passos,
  D.~Cournapeau, M.~Brucher, M.~Perrot, and E.~Duchesnay.
\newblock Scikit-learn: Machine learning in {P}ython.
\newblock \emph{Journal of Machine Learning Research}, 12:\penalty0 2825--2830,
  2011.
\newblock \doi{https://doi.org/10.5555/1953048.2078195}.

\bibitem[Perry et~al.(2021)Perry, Mischler, Guo, Lee, Chang, Koul, Franz,
  Richard, Carmichael, Ablin, Gramfort, and Vogelstein]{perry2021mvlearn}
R.~Perry, G.~Mischler, R.~Guo, T.~Lee, A.~Chang, A.~Koul, C.~Franz, H.~Richard,
  I.~Carmichael, P.~Ablin, A.~Gramfort, and J.~T. Vogelstein.
\newblock mvlearn: Multiview machine learning in python.
\newblock \emph{Journal of Machine Learning Research}, 22\penalty0
  (109):\penalty0 1--7, 2021.
\newblock \doi{https://doi.org/10.5555/3546258.3546367}.

\bibitem[Saeys et~al.(2007)Saeys, Inza, and Larrañaga]{saeys2007}
Y.~Saeys, I.~Inza, and P.~Larrañaga.
\newblock {A review of feature selection techniques in bioinformatics}.
\newblock \emph{Bioinformatics (Oxford, England)}, 23\penalty0 (19):\penalty0
  2507--2517, 08 2007.
\newblock \doi{https://doi.org/10.1093/bioinformatics/btm344}.

\bibitem[Shim and Kang(2022)]{SHIM2022}
J.~Shim and S.~Kang.
\newblock Domain-adaptive active learning for cost-effective virtual metrology
  modeling.
\newblock \emph{Computers in Industry}, 135, 2022.
\newblock \doi{https://doi.org/10.1016/j.compind.2021.103572}.

\bibitem[Su et~al.(2007)Su, Jeng, Huang, Yu, Hung, and Chao]{su2007control}
A.-J. Su, J.-C. Jeng, H.-P. Huang, C.-C. Yu, S.-Y. Hung, and C.-K. Chao.
\newblock Control relevant issues in semiconductor manufacturing: Overview with
  some new results.
\newblock \emph{Control Engineering Practice}, 15\penalty0 (10):\penalty0
  1268--1279, 2007.
\newblock \doi{https://doi.org/10.1016/j.conengprac.2006.11.003}.
\newblock Special Issue - International Symposium on Advanced Control of
  Chemical Processes (ADCHEM).

\bibitem[Sun et~al.(2016)Sun, Feng, and Saenko]{sun2016return}
B.~Sun, J.~Feng, and K.~Saenko.
\newblock Return of frustratingly easy domain adaptation.
\newblock In \emph{Proceedings of the AAAI Conference on Artificial
  Intelligence}, volume~30, 2016.
\newblock \doi{https://doi.org/10.5555/3016100.3016186}.

\bibitem[Sun(2013)]{sun2013mvl}
S.~Sun.
\newblock A survey of multi-view machine learning.
\newblock \emph{Neural Computing and Applications}, 23, 12 2013.
\newblock \doi{https://doi.org/10.1007/s00521-013-1362-6}.

\bibitem[Susto and Beghi(2012)]{susto2012least}
G.~A. Susto and A.~Beghi.
\newblock Least angle regression for semiconductor manufacturing modeling.
\newblock In \emph{2012 IEEE International Conference on Control Applications},
  pages 658--663. IEEE, 2012.
\newblock \doi{https://doi.org/10.1109/CCA.2012.6402409}.

\bibitem[Susto et~al.(2015)Susto, Schirru, Pampuri, McLoone, and
  Beghi]{susto2015pdm}
G.~A. Susto, A.~Schirru, S.~Pampuri, S.~McLoone, and A.~Beghi.
\newblock Machine learning for predictive maintenance: A multiple classifier
  approach.
\newblock \emph{IEEE Transactions on Industrial Informatics}, 11\penalty0
  (3):\penalty0 812--820, 2015.
\newblock \doi{https://doi.org/10.1109/TII.2014.2349359}.

\bibitem[Taigman et~al.(2017)Taigman, Polyak, and Wolf]{taigman2017}
Y.~Taigman, A.~Polyak, and L.~Wolf.
\newblock Unsupervised cross-domain image generation.
\newblock \emph{ArXiv}, abs/1611.02200, 2017.
\newblock \doi{https://doi.org/10.48550/arxiv.1611.02200}.

\bibitem[Tsai et~al.(2016)Tsai, Yeh, and Wang]{tsai2016hetero}
Y.-H.~H. Tsai, Y.-R. Yeh, and Y.-C.~F. Wang.
\newblock Heterogeneous domain adaptation with label and structure consistency.
\newblock In \emph{2016 IEEE International Conference on Acoustics, Speech and
  Signal Processing (ICASSP)}, pages 2842--2846, 2016.
\newblock \doi{https://doi.org/10.1007/s11042-020-08731-x}.

\bibitem[Tsutsui and Matsuzawa(2019)]{tsutsui2019virtual}
T.~Tsutsui and T.~Matsuzawa.
\newblock Virtual metrology model robustness against chamber condition
  variation using deep learning.
\newblock \emph{IEEE Transactions on Semiconductor Manufacturing}, 32\penalty0
  (4):\penalty0 428--433, 2019.
\newblock \doi{https://doi.org/10.1109/ISSM.2018.8651170}.

\bibitem[Vincent et~al.(2020)Vincent, Wannes, and Jesse]{transfertools}
V.~Vincent, M.~Wannes, and D.~Jesse.
\newblock Transfer learning for anomaly detection through localized and
  unsupervised instance selection.
\newblock \emph{Proceedings of the AAAI Conference on Artificial Intelligence},
  34\penalty0 (04):\penalty0 6054--6061, Apr. 2020.
\newblock \doi{https://doi.org/10.1609/aaai.v34i04.6068}.

\bibitem[Wang and Deng(2018)]{wang2018deep}
M.~Wang and W.~Deng.
\newblock Deep visual domain adaptation: A survey.
\newblock \emph{Neurocomputing}, 312:\penalty0 135--153, 2018.
\newblock \doi{https://doi.org/10.1016/j.neucom.2018.05.083}.

\bibitem[Wang et~al.(2004)Wang, Bovik, Sheikh, and Simoncelli]{ssim}
Z.~Wang, A.~Bovik, H.~Sheikh, and E.~Simoncelli.
\newblock Image quality assessment: from error visibility to structural
  similarity.
\newblock \emph{IEEE Transactions on Image Processing}, 13\penalty0
  (4):\penalty0 600--612, 2004.
\newblock \doi{https://doi.org/10.1109/TIP.2003.819861}.

\bibitem[Xu et~al.(2013)Xu, Tao, and Xu]{multiview}
C.~Xu, D.~Tao, and C.~Xu.
\newblock A survey on multi-view learning.
\newblock \emph{arXiv}, 2013.
\newblock \doi{https://doi.org/10.48550/ARXIV.1304.5634}.

\bibitem[Yan et~al.(2021)Yan, Hu, Mao, Ye, and Yu]{YAN2021106}
X.~Yan, S.~Hu, Y.~Mao, Y.~Ye, and H.~Yu.
\newblock Deep multi-view learning methods: A review.
\newblock \emph{Neurocomputing}, 448:\penalty0 106--129, 2021.
\newblock \doi{https://doi.org/10.1016/j.neucom.2021.03.090}.

\bibitem[Yang et~al.(2020)Yang, Song, Jin, and Du]{ijcai2020}
S.~Yang, G.~Song, Y.~Jin, and L.~Du.
\newblock Domain adaptive classification on heterogeneous information networks.
\newblock In C.~Bessiere, editor, \emph{Proceedings of the Twenty-Ninth
  International Joint Conference on Artificial Intelligence, {IJCAI-20}}, pages
  1410--1416. International Joint Conferences on Artificial Intelligence
  Organization, 7 2020.
\newblock \doi{https://doi.org/10.5555/3491440.3491636}.

\bibitem[Yu et~al.(2021)Yu, Li, Zhao, Liu, and Lin]{yucca}
Q.~Yu, L.~Li, H.~Zhao, Y.~Liu, and K.-Y. Lin.
\newblock Evaluation system and correlation analysis for determining the
  performance of a semiconductor manufacturing system.
\newblock \emph{Complex System Modeling and Simulation}, 1\penalty0
  (3):\penalty0 218--231, 2021.
\newblock \doi{https://doi.org/10.23919/CSMS.2021.0015}.

\bibitem[Zhu et~al.(2017)Zhu, Park, Isola, and Efros]{zhu2017}
J.-Y. Zhu, T.~Park, P.~Isola, and A.~A. Efros.
\newblock Unpaired image-to-image translation using cycle-consistent
  adversarial networks.
\newblock In \emph{2017 IEEE International Conference on Computer Vision
  (ICCV)}, pages 2242--2251, 2017.
\newblock \doi{10.1109/ICCV.2017.244}.

\end{thebibliography}






\end{document}